\pdfoutput=1

\documentclass[11pt]{article}

\usepackage{acl}

\usepackage{times}
\usepackage{latexsym}

\usepackage[T1]{fontenc}

\usepackage[utf8]{inputenc}

\usepackage{microtype}

\usepackage{inconsolata}

\usepackage{natbib} 
\usepackage{booktabs}
\usepackage{comment}
\usepackage{graphicx}
\usepackage{subfig}
\usepackage{fancyhdr}
\usepackage{amsmath}
\usepackage{amssymb}
\usepackage{xurl}
\usepackage{adjustbox}
\usepackage{arydshln}
\usepackage{relsize}
\usepackage{wasysym}
\usepackage{multirow}
\usepackage[makeroom]{cancel}
\usepackage{scalerel,graphicx,xparse}
\usepackage{hyperref}
\usepackage{xcolor}
\usepackage{algpseudocode}
\usepackage{lscape}

\usepackage{tikz}
\usetikzlibrary{positioning, arrows.meta, calc,shapes.misc, fit}

\title{CMA-R: Causal Mediation Analysis for Explaining Rumour Detection}

\author{Lin Tian \\ RMIT University, Australia \\ lin.tian2@student.rmit.edu.au
         \And
         Xiuzhen Zhang \\ RMIT University, Australia \\ xiuzhen.zhang@rmit.edu.au \AND
         Jey Han Lau \\ The University of Melbourne, Australia \\ jeyhan.lau@gmail.com}
         
\begin{document}
\maketitle
\begin{abstract}
We apply causal mediation analysis to explain the decision-making process of neural models for rumour detection on Twitter.
Interventions at the input and network level reveal the causal impacts of tweets and words in the model output.
We find that our approach CMA-R -- Causal Mediation Analysis for Rumour detection -- identifies salient tweets that explain model predictions and show strong agreement with human judgements for critical tweets determining the truthfulness of stories.
CMA-R can further highlight causally impactful words in the salient tweets, providing another layer of interpretability and transparency into these blackbox rumour detection systems. Code is available at: \url{https://github.com/ltian678/cma-r}.

\end{abstract}

\section{Introduction}

There has been substantial work on understanding the inner workings of neural models via attention mechanisms~\citep{clark2019does}, local surrogated approaches~\citep{ribeiro2016should,lundberg2017unified,kokalj2021bert} or integrated gradient based methods~\citep{sundararajan2017axiomatic}.
Existing works on explainable fake news or rumour detection by and large use attention weights to explain model decision \citep{shu2019defend,khoo2020interpretable,lu2020gcan,li2021meet}, but \citet{pruthi+:2020} found that the use of attention as explanation is problematic: removing words with high attention appears to have little effect on the final prediction, suggesting that attention doesn't explain the decision process.

To address these limitations,
in this paper, we propose CMA-R -- Causal Mediation Analysis for Rumour detection -- grounded in causal mediation analysis (CMA~\citep{pearl2001direct},  
as illustrated in Figure~\ref{fig:causal_def}) to interpret decisions for rumour detection models. 
CMA-R is a significant departure from existing interpretation methods, as it provides greater explanatory power from assessing causal relations instead of correlations. 
Different from studies ~\citep{vig2020investigating} that apply CMA to examine the causal structure from network components to predictions, we perform intervention in the input and network to determine the tweets and words that are \textit{causally implicated} in the final prediction and verify them with human expert annotations. 
Using a rumour dataset that has been annotated by journalists to highlight critical tweets that determine the truthfulness of a story, we assess the salient tweets extracted by CMA-R and other interpretation methods  (e.g.\ attention) and found that CMA-R yields better alignment with human judgements, empirically demonstrating that it is important to consider causality for explaining model decisions. CMA-R also allows us to highlight impactful words in those salient tweets, providing another mechanism to interpret rumour detection models.

The main contributions of this work are as follows:
\begin{itemize}
    \item CMA-R is a novel application on interpreting rumour detection systems model decisions by performing interventions in the input and network that aims to identify tweets and words causally implicated in the final prediction.
    \item CMA-R can highlight impactful words in salient tweets via neuron level interventions, providing a refined mechanism for interpreting rumour detection models.
    \item Our findings show that CMA-R aligns more closely with human judgments on a journalist-annotated rumour dataset.
\end{itemize}

\section{Related Work}
We briefly summarise prior studies from three related areas: explainable artificial intelligence, causal mediation analysis and rumour detection.

Explainable artificial intelligence aims to create a suite of techniques to produce interpretable artificial intelligence systems, which are often driven by deep learning~\citep{gunning2019xai}.
Broadly speaking there are two approaches: model-agnostic and model-specific methods.
Model-agnostic approaches such as LIME (Local Interpretable Model-Agnostic Explanations)~\cite{ribeiro2016should} and SHAP (SHapley Additive exPlanations)~\cite{lundberg2017unified,kokalj2021bert} build local surrogate models to approximate the predictions of the original model.
Model-specific techniques use feature visualisation~\citep{vig2019bertviz} and attention mechanisms~\cite{clark2019does} to explain the decision-making process.
Additionally, rationalisation-based approaches focus on generating textual explanations that rationalise a model's decision. 
The explanations mimic human reasoning and provide narrative or rationale for why a model made a certain decision~\citep{rajani2019explain,pan2022accurate,liu2022fr,liuetal2023mgr,chrysostomou2022flexible}.
It is not a way to explain a model's internal decision-making processes, but a method for rationalising the behaviour and justifying its predictions.

Causal mediation analysis (CMA) aims to uncover cause-and-effect relationships, and its application to understanding deep learning models is emerging ~\citep{vig2020investigating,feder2022causal,qian2021my}. CMA-R goes beyond understanding the correlations between the input and output, but instead attempts to the causal structure for model decisions.
In this paper, we employ CMA-R to understand how intervention at both the word and neuron levels affect the model's predictions.

Deep learning is the dominant approach for automatic detection of online rumours and fake news~\citep{shu2019defend,khoo2020interpretable,lu2020gcan,li2021meet}.
Attention mechanisms have been widely used to explain model decisions~\citep{shu2019defend,khoo2020interpretable,lu2020gcan},
but there is emerging evidence showing that correlation does not always constitute explanation \citep{jain2019attention,serrano2019attention,pruthi+:2020}.



\section{Preliminaries}
\label{sec:problem-statement}
\begin{figure}[t]
\centering
  \includegraphics[width=\linewidth]{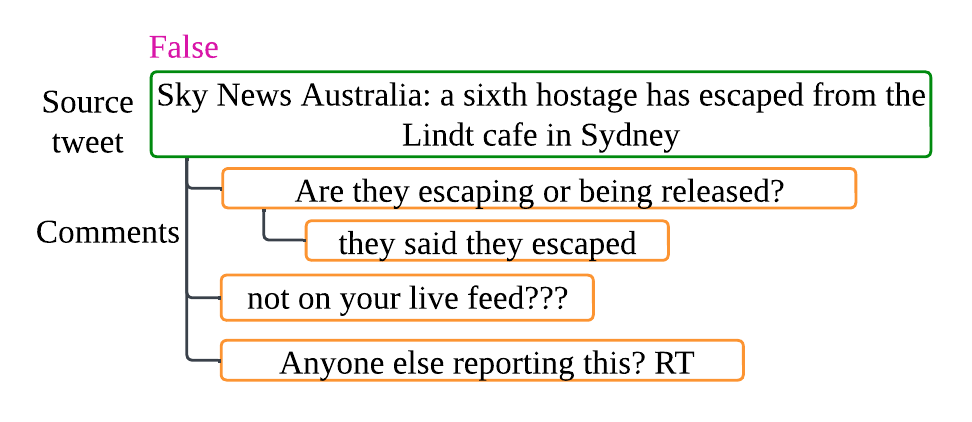}
  \caption{Labelled source tweet in PHEME.}
  \label{fig:source-pheme}
\end{figure}

\begin{figure}[t]
\centering
  \includegraphics[width=\linewidth]{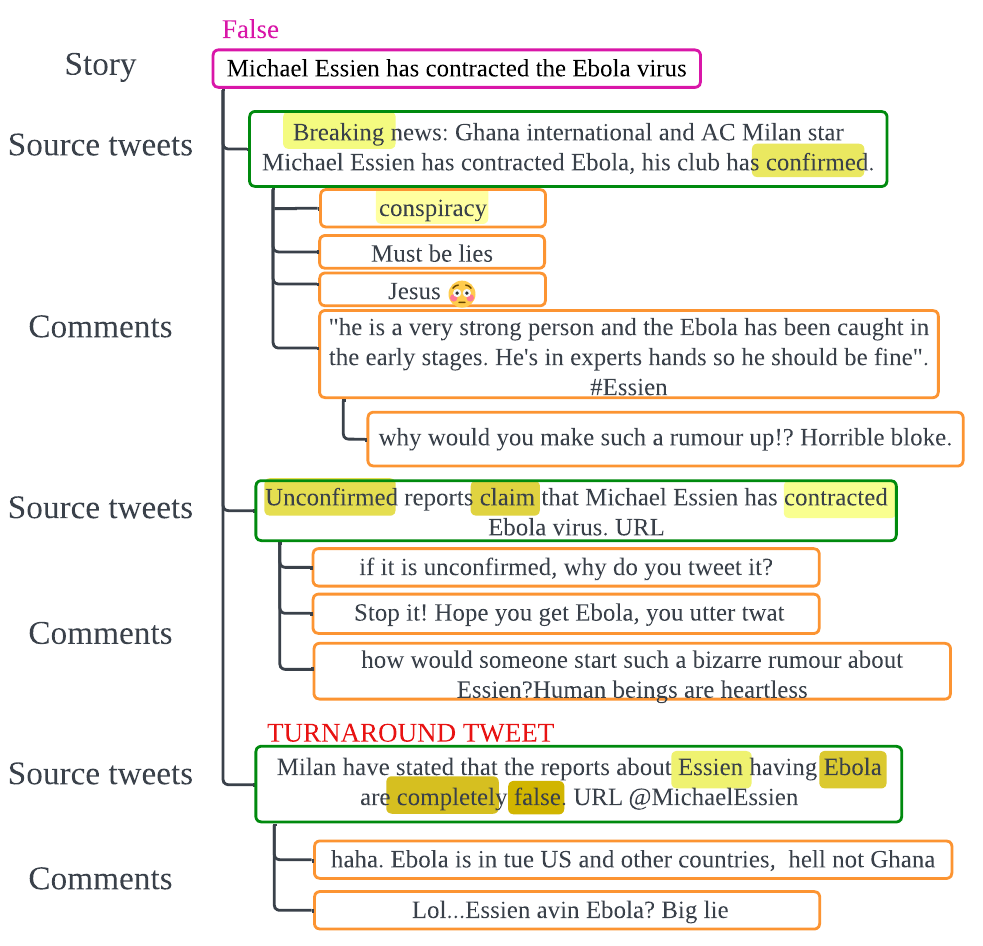}
  \caption{A labelled story in PHEME. Additional stories can be found in Appendix~\ref{sec: labelled_samples_appendix}. }
  \label{fig:story-pheme}
\end{figure}

Let $X = \{x_0,x_1,x_2,...,x_n\} $ be a set of events, where an event $x_i$ consists of either: (1) a source tweet and its comments (Figure \ref{fig:source-pheme}); or (2) a story with a set of source tweets and their comments (Figure \ref{fig:story-pheme}). 
Each event $x_i$ is associated with a rumour label $y_i\in Y$, 
where $Y$ represents three rumour veracity classes (true, false or unverified).
A rumour detection system is trained (with labelled data) to learn $f:X\rightarrow Y$.


\begin{figure}[t]
\centering
\begin{tikzpicture}[node distance=2cm, thick,
    circleNode/.style={circle, draw, minimum size=0.8cm, align=center},
    arrowStyle/.style={->, >={Latex[length=3mm, width=2mm]}}
]

\node[circleNode] (X) {X};
\node[circleNode, right=4.5cm of X] (Y) {Y}; 
\node[circleNode] (Z) at ($(X)+(2.5cm,1.25cm)$) {Z}; 

\draw[arrowStyle] (X) -- (Y) node[midway, below] {Direct Effect};
\draw[arrowStyle, dashed] (Z) -- (Y) node[bend right=30] [midway, above right] {Indirect Effect};
\draw[arrowStyle, dashed] (X) -- (Z);

\end{tikzpicture}
\caption{Casual mediation analysis.}
\label{fig:causal_def}
\end{figure}
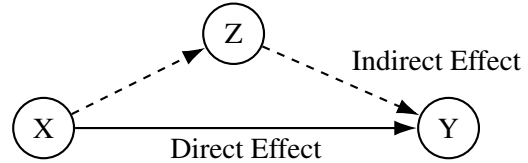

\begin{figure*}[t]
\centering
  \includegraphics[width=\linewidth]{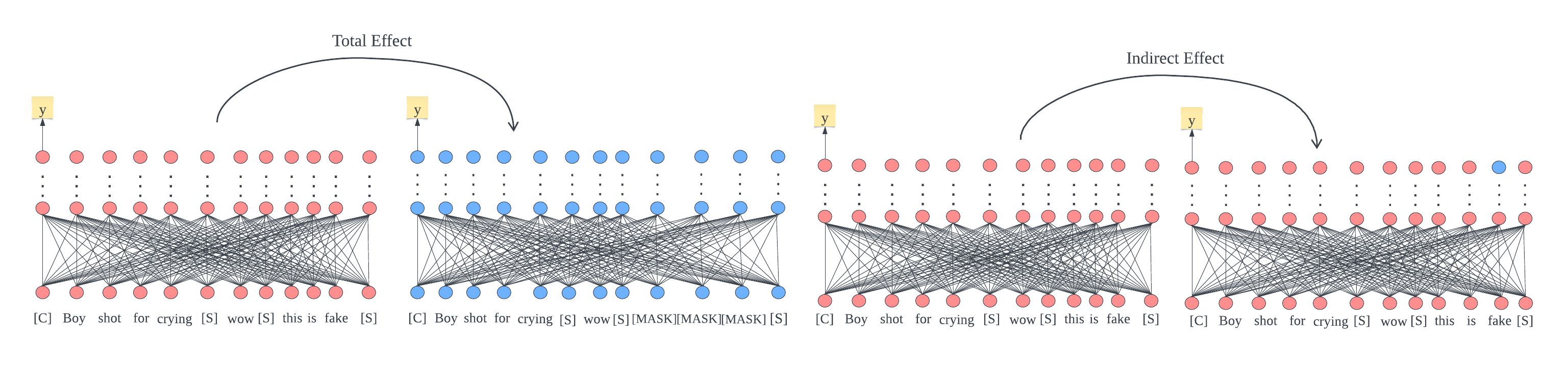}
  \caption{Total effect and indirect effect in CMA-R. [C] ([CLS]) and [S] ([SEP]) represent special tokens.}
  \label{fig:causal_all}
\end{figure*}

\section{Methodology}

CMA-R allows us to analyse the change of a response variable ($y$) following a treatment ($x$) --- e.g.\ in the biomedical domain this could mean the patient's health outcome given a treatment --- and it does so by considering \textit{mediators} ($z$), intermediate factors that produce an \textit{indirect effect}. 
As shown in Figure~\ref{fig:causal_def}, a mediator ($z$) is added to take into account its indirect effect. \citet{vig2020investigating} introduce CMA as a means to explain the decision of a neural model, by viewing the model input as $x$, the model output (decision) as $y$, and the neurons in the model as $z$.
In CMA-R, $x$ represents an event and $y$ a rumour label, and the tweets in $x$ are encoded using a sequence network (e.g.\ BERT \cite{devlin2018bert}). The tweets in $x$ may be concatenated as a string or represented as a graphs (to model the conversation structure), depending on the rumour detection model (Section \ref{sec:models}).



\subsection{Total Effects}
\label{sec:total-effects}

To measure the causal impact of a tweet (or a set of tweets) in an event ($x$) that contribute to a model prediction ($y$), we can perform intervention by masking it out and computing the total effect:
\begin{equation}
\text{TE} = D(\mathbf{y}_{\text{null}}(x), \mathbf{y}_{\text{mask-text}}(x))
\end{equation}
where ``null'' and ``mask-text''  denote the intervention operations:  the former performs no intervention and the latter masks out tweet(s) in the input  (Figure \ref{fig:causal_all} left); $\mathbf{y}$ represents the output probability distribution over the three veracity classes and $D$ is a distance metric between two probability distributions (Section \ref{sec:distance-metric}).


\subsection{Indirect Effects}
\label{sec:indirect-effects}

CMA-R also allows us to measure the causal impact of a neuron (or a set of neurons) by computing the indirect effect. The idea is to replace the value of a neuron in the pre-intervention network using that of the post-intervention network and measure how much that changes prediction. Formally:
\begin{equation}
\text{IE} = D(\mathbf{y}_{\text{null}}(x), \mathbf{y}_{\text{replace-neuron}}(x))
\end{equation}
where ``replace-neuron'' is the intervention operation for neuron replacement (Figure \ref{fig:causal_all} right).
Given that we use sequence networks (e.g.\ recurrent or transformer) to encode text, we can target neurons associated with words to measure the causal impact of each word, e.g.\ for a transformer encoder we can perform this replacement for neurons at different transformer layers that correspond to a word.




\subsection{Distance Metric}
\label{sec:distance-metric}

\citet{vig2020investigating} use CMA for a task which has a binary outcome, and they propose computing the ratio between the probabilities of the positive class pre- and post-intervention to compute total/indirect effect. In our case (CMA-R), as we are dealing with a multi-class classification problem (3 veracity classes), we experiment with the following two distance metrics for two probability distributions \cite{dwork2012fairness}:
\begin{align*}
T_1 &= \frac{1}{2} \sum_{y \in Y}|y_{\text{null}}(x) - y_{\text{intervention}}(x)| \\
T_2 &= e^{\max_{y \in Y}\log (  \max (r_y, 1/r_y))}
\end{align*}
where $y_{\text{null}}(x)$ and $y_{\text{intervention}}(x)$ denote the output probability of a label without and with intervention respectively and $r_y$ = $\frac{y_{\text{null}}(x)}{y_{\text{intervention}}(x)}$.
To rank the causal impact of tweets (total effect), we compute two rankings using the two distance metrics and sum the rankings to produce the final ranking. We rank the causal impacts of words (indirect effect) in the same way (i.e. via sum rank).

\section{Experiment}

\subsection{Datasets}

We use two variants of PHEME that contain veracity labels at two different levels: (1) source tweet (Figure \ref{fig:source-pheme}; \citet{kochkina2018all});\footnote{\url{figshare.com/articles/dataset/PHEME_dataset_for_Rumour_Detection_and_Veracity_Classification/6392078}} and (2) story (Figure \ref{fig:story-pheme}; \citet{zubiaga2016analysing}).\footnote{\url{figshare.com/articles/dataset/PHEME_rumour_scheme_dataset_journalism_use_case/2068650}} The former contains 29,387 labelled source tweets (with comments) while the latter has 46 labelled stories (each story can be interpreted as a news event that is linked to a number of related source tweets).\footnote{The description of a story, e.g.\ \textit{Michael Esseien has contracted the Ebola virus} in Figure \ref{fig:story-pheme} is written by journalists.}
Each labelled story however, is also annotated with a ``turnaround tweet'' -- the source tweet judged (by journalists) to be the critical tweet that determined the final veracity of a story.\footnote{Technically, original dataset has 240 labelled stories, but only 46 of them has a turnaround tweet.} We use the (larger) first PHEME variant to train a rumour classifier, and then apply the trained classifier to the (smaller) second PHEME variant to classify the stories and assess whether the salient source tweets extracted by CMA-R correspond to the ground truth turnaround tweets. Note that there is no overlap in terms of source tweets between the first and second PHEME variant, and so the rumour classifier has not ``seen'' any of the stories.


\subsection{Models and Training Strategies}
\label{sec:models}

We experiment with three models with different architecture for encoding the tweets in $x$: (1) \textbf{one-tier transformer} uses RoBERTa \cite{liu2019roberta} to encode the tweets concatenated as a string; (2) \textbf{two-tier transformer} \cite{tian2022duck} uses BERT \cite{devlin2018bert} to encode each tweet separately and then another (randomly initialised) transformer to encode the sequence of [CLS] output embeddings from BERT; and (3) \textbf{DUCK} \cite{tian2022duck} uses BERT to encode each pair of parent-child\footnote{Child tweet here means a replying comment.} tweet and a graph attention network to encode the output from BERT to capture the conversation structure.\footnote{In the original paper the best DUCK variant is an ensemble that combines all three architectures.} DUCK represents the current state-of-the-art for rumour detection.

In terms of training strategy, we explore two methods: (1) fine-tune using PHEME; and (2) fine-tune using Twitter15/16 and PHEME (in sequence). As Twitter15/16 is a larger labelled rumour dataset, we suspect the additional training would improve the models' veracity prediction performance.

\subsection{Baseline Interpretation Models}

We test CMA-R with three other common baselines to extract salient tweets: (1) \textbf{attention}: we aggregate the attention weights for each word (one/two-tier transformer) or node (DUCK) and then rank each source tweet$+$comments by computing the average attention weight over their words (one/two-tier transformer) or nodes (DUCK); (2) \textbf{local}: we use LIME \cite{ribeiro2016should} to compute word weights, and aggregate word weights in the same way as described before;\footnote{We use the following code for one/two-tier transformer and DUCK respectively: \url{https://github.com/cdpierse/transformers-interpret}, \url{https://github.com/mims-harvard/GraphXAI}.}; (3) \textbf{gradient}: we compute word weights based on their gradients \cite{sundararajan2017axiomatic} and aggregate word weights.

We further compare with three baseline systems for explainable fake news and rumour detection: (1) dEFEND \citep{shu2019defend} generates attention scores for both source tweets and their comments. The comment receiving the highest attention score is selected as the ``turnaround tweet'' – the key tweet that provides the most explanatory power in the context of a rumour. (2) GCAN \citep{lu2020gcan} does not explicitly identify the most explainable tweet in its original formulation. Attention scores are generated through its post and propagation attention mechanism. We adapted this by selecting tweets with the highest attention scores in this mechanism, assuming these to be the most relevant for explanation purposes. (3) StA-HiTPLAN \citep{khoo2020interpretable} provides post-level explanations based on the attention scores of the last layer. We used these post-level explanations to match back to the human-identified decision points in our datasets, assuming that higher attention scores correlate with greater explanatory relevance. All three baselines belong to attention-based approaches.
\begin{table} [tbh!]
\begin{center}
\begin{adjustbox}{max width=\linewidth}
\begin{tabular}
{l@{\;\;}cccccc}
\toprule
\toprule
\multirow{2}{*}{Model} &\multirow{2}{*}{F1} & \multicolumn{5}{c}{{Turnaround Accuracy}} \\
\cline{3-7}
 & & R & A & L & G & C \\
\midrule
\multicolumn{2}{l}{Fine-tune with PHEME} \\
\midrule
One-Tier & 0.70 & 0.05 &0.26 &0.20 &0.33 &0.41*  \\
Two-Tier & 0.73 &0.05 &0.28 &0.28 &0.41 &0.54* \\
DUCK & 0.81 &0.05 &0.26 &0.26 &0.46 &0.65* \\
\midrule
dEFEND~\citep{shu2019defend} &0.62 & - & 0.20 & - & - & - \\
GCAN~\citep{lu2020gcan} &0.72 &- &0.28 & - & - & - \\
StA-HiTPLAN~\citep{khoo2020interpretable} &0.39 &- &0.09 & - & - & - \\
\midrule
\multicolumn{2}{l}{Fine-tune with Twitter15/16 and PHEME} \\
\midrule
One-tier &0.72 &0.05 &0.26 &0.20 &0.37 &0.43*\\
Two-tier &0.75 &0.05 &0.30 &0.28 &0.43 &0.61* \\
DUCK &0.85 &0.05 &0.30 &0.28 &0.48 &0.70* \\
\midrule
dEFEND~\citep{shu2019defend} &0.66 & - & 0.22 & - & - & - \\
GCAN~\citep{lu2020gcan} &0.75 &- &0.28 & - & - & - \\
StA-HiTPLAN~\citep{khoo2020interpretable} &0.42 &- &0.09 & - & - & - \\
\bottomrule
\bottomrule
\end{tabular}
\end{adjustbox}
\end{center}
\caption{Turnaround accuracy results. F1 denotes rumour classification performance. R: random baseline; A: attention; L: local; G: gradient; and C: CMA-R. An asterisk (*) indicates that the result is statistically significant with \( p \ll 0.05 \). Detailed scores are in Appendix~\ref{sec:statisticaltest_appendix}.}
\label{tab:turnaround_res}	
\end{table}

\begin{figure}[t]
\centering
  \includegraphics[width=\linewidth]{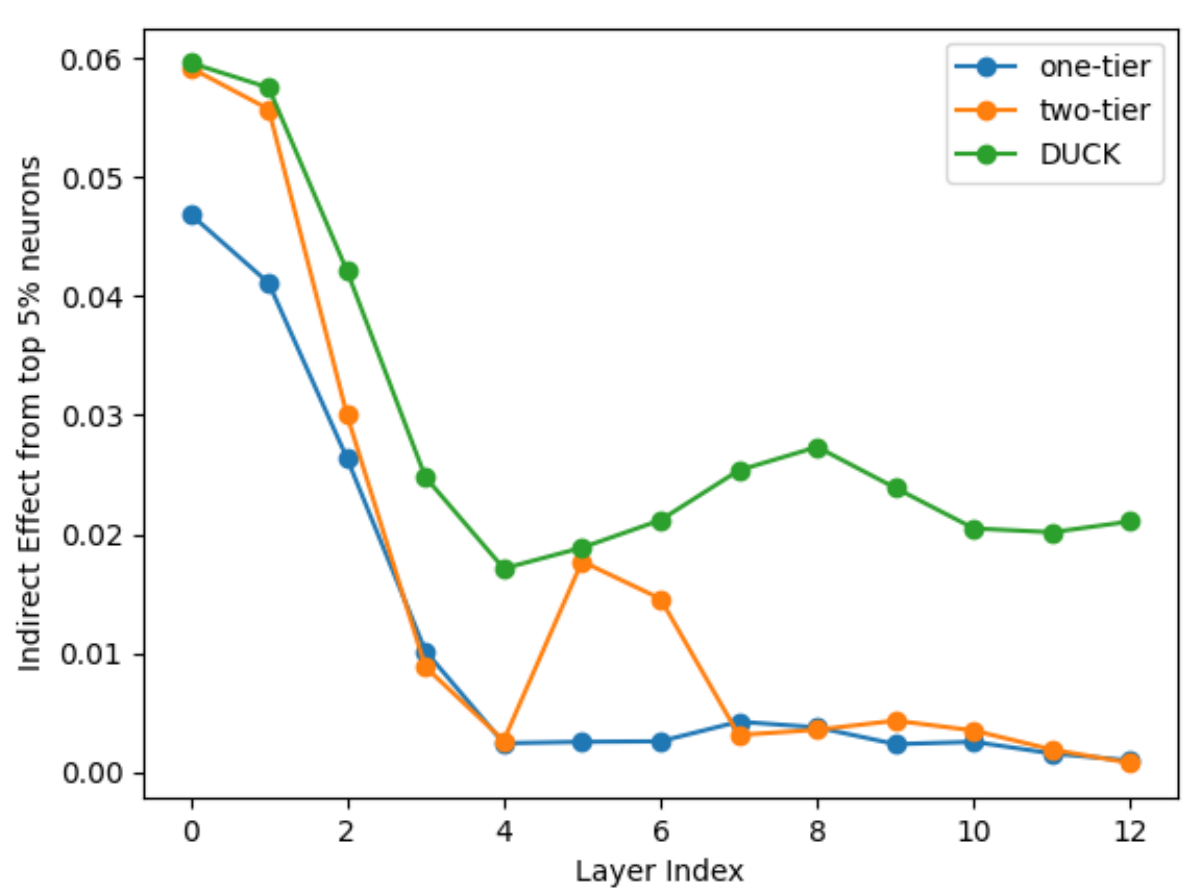}
  \caption{Indirect effects over different layers}
  \label{fig:layer_res}
\end{figure}

\section{Results}
\subsection{Turnaround Accuracy}

We now assess how well the different interpretation methods pick up the correct turnaround tweets. Note that for CMA-R, when performing the ``mask-text'' intervention (Section \ref{sec:total-effects}) we mask each source tweet (and their associated comments) one at a time in order to determine which source tweet has the most causal impact.
Table~\ref{tab:turnaround_res} presents the results. ``R'' denotes a random baseline where a random source tweet is chosen; 0.05 indicates on average 20 source tweets in a story. It is therefore a non-trivial task to identify the turnaround tweet.

We first look at the two fine-tuning strategies, and we see (without surprise) that the use of additional training data (Twitter15/16) improves rumour detection performance for all models, and that in turn leads to higher turnaround accuracy. Comparing the three models, DUCK is the clear winner here. Looking at the different interpretability methods (attention, local, gradient and CMA-R), we have a consistent observation: CMA-R is much more accurate at extracting the correct turnaround tweets, followed by gradient. 
Compared with existing explainable rumour detection approaches~\citep{shu2019defend,lu2020gcan,khoo2020interpretable}, we still can see that CMA-R better aligns with the human decision points.
At a higher level, these results imply that it is important that we consider causal relations rather than correlations when interpreting model decisions.\footnote{In Appendix~\ref{sec:turnaround_appendix}, we provide further analyses where we consider only stories where a model have predicted the rumour veracity correctly (true or false). The general finding is broadly the same, where DUCK$+$CMA-R is the best combination in terms of veracity and turnaround prediction.} We next present additional analyses, and in these experiments we use Twitter15/16 and PHEME fine-tuned DUCK.



\subsection{Salient Words}

We use CMA-R to extract the most salient words by computing the indirect effects. When performing the ``replace-neuron'' intervention (Section \ref{sec:indirect-effects}), we replace the neurons for one transformer layer at a time, word by word. As such, we have a ranking of words for each layer, and we sum the rankings from the word embeddings and first six transformer layers. We highlight (in yellow) the most impactful words for a story in Figure \ref{fig:story-pheme}. Interestingly, CMA-R extracts a number of intuitively critical words in the turnaround tweet, suggesting that it is focusing on the right words when making its decision.



\subsection{Sparsity and Layer effects distribution}

Following \citet{vig2020investigating} we also compute the indirect effects of the top neurons in different layers; results in 
Figure~\ref{fig:layer_res}.
In terms of the magnitude of indirect effects, DUCK seem to produce substantially higher effects. Across the layers, the earlier layers appear to have a much larger impact (this isn't a surprising finding, as they are connected to more neurons in the network). Interestingly, though, we see a small bump in the middle layers of DUCK and two-tier transformer, which \citet{vig2020investigating} also found. In Appendix~\ref{sec:totaleffects_appendix}, we present further analyses on the total effects.



\section{Conclusion}

We employed causal mediation analysis to understand the inner workings of rumour detection models. By performing interventions at the input and network levels, we show that our approach CMA-R can find tweets and words having the most causal impact for model decisions. To evaluate the ``quality'' of these insights, we train rumour detection models of differing complexity and
compare CMA-R to current interpretation methods to assess how well the extracted salient tweets align with human judgements. 
Empirical results demonstrate that CMA-R is consistently the best method, suggesting that causal relations, rather than correlations, can better interpret model decisions. CMA-R provides further mechanism to hone in on the words for the most causal impact, and qualitative analysis reveals that the best rumour detection model is focusing on intuitively important words when determining the veracity of a story.


\section{Limitations}
We acknowledge that the size of our test data (story-annotated PHEME) is relatively small (46 instances), and this points to the laborious and difficult nature of the annotation task. That said, we contend that our results constitute one of the first studies in rumour detection that attempts to empirically validate the quality of insights produced by interpretation methods. 
To ensure the robustness of our results, we have conducted significance tests (results included in Appendix~\ref{sec:statisticaltest_appendix}).

While our work primarily focuses on applying causal mediation analysis to text-based rumour detection models, it is important to acknowledge that we did not apply user-based or propagation-based interventions in this particular study. However, the emphasis on text-based analysis provides a foundation for future investigations that can extend our methodology to encompass other methods and incorporate a more comprehensive understanding of rumour detection systems.

\section*{Acknowledgement}
This research is supported in part by the Australian Research Council Discovery Project DP200101441. 
Lin Tian is supported by the RMIT University Vice-Chancellor PhD Scholarship (VCPS).

\bibliography{custom}

\appendix

\section{Magnitude of Total Effects}
\label{sec:totaleffects_appendix}

\begin{table}[tbh!]
\begin{center}
\begin{adjustbox}{max width=\linewidth}
\begin{tabular}{rccccc}
\toprule
\toprule
& Model   & Params    & $T_1$ & $T_2$  \\ \toprule
& One-tier &125M  &0.27 &0.12 \\
& Two-tier &165M &0.30 &0.19 \\
& DUCK &143M  &0.73 &0.55 \\
\bottomrule
\bottomrule 
\end{tabular}
\end{adjustbox}
\end{center}
\caption{\label{tab:te_analysis} Average Total Effects.}
\end{table}

To calculate the total effect for each model, we compute the average total effects by aggregating the individual effects across all 46 test instances. These effects represent the cumulative influence of the model neurons on the interventions. 
Table~\ref{tab:te_analysis} shows the magnitude of average total effects (over source tweets and stories) for the two distance metrics. Interestingly, we find that the total effects using DUCK appears to be subtantially larger.

\begin{table*}
\begin{center}
\begin{adjustbox}{max width=\linewidth}
\begin{tabular}
{lcc@{\;\;}c@{\;\;}c@{\;\;}c@{\;\;}cc@{\;\;}c@{\;\;}c@{\;\;}c@{\;\;}c}
\toprule
\toprule
\multirow{2}{*}{Model} &\multirow{2}{*}{F1} & \multicolumn{5}{c}{Conditional TRUE (27)} & \multicolumn{5}{c}{Conditional FALSE (19)} \\
\cline{3-12}
 &  &\#TP & Attention & Local & IG & CMA-R &\#TP & Attention & Local & IG & CMA-R \\
\midrule
\multicolumn{2}{l}{Fine-tune with PHEME} \\
\midrule
One-Tier & 0.70  &17 &0.18 &0.24 &0.24 &0.24 &11 &0.09 &0 &0.36 &0.64 \\
Two-Tier & 0.73 &18 &0.22 &0.22 &0.28 &0.33 &12 &0.08 &0.17 &0.42 &0.58 \\
DUCK & 0.81 &23 &0.26 &0.35 &0.43 &0.52 &14 &0.14 &0.29 &0.50 &0.57 \\
\midrule
\multicolumn{2}{l}{Fine-tune with Twitter15/16 and PHEME} \\
\midrule
One-tier &0.72 &20 &0.20 &0.10 &0.20 &0.30 &12 &0.17 &0.08 &0.33 &0.58  \\
Two-tier &0.75 &21 &0.19 &0.29 &0.48 &0.57 &13 &0.08 &0.23 &0.46 &0.62 \\
DUCK &0.85 &23 &0.35 &0.35 &0.52 &0.61 &15 &0.13 &0.27 &0.47 &0.60  \\
\bottomrule
\bottomrule
\end{tabular}
\end{adjustbox}
\end{center}
\caption{Turnaround accuracy results. F1 denotes rumour classification performance. \#TP represents the number of correct classified instances.}
\label{tab:turnaround_res_appendix}	
\end{table*}

\section{Turnaround Accuracy}
\label{sec:turnaround_appendix}
To better understand the effectiveness of causal mediation analysis as a way to explain model decisions, we further measure its performance under the conditional scenario. In this case, we do care about whether the model correctly predicted the rumour's truthfulness. Since resolving tweets lead to a rumour being labelled as true or false, we can measure how accurately the model predicts this. In this scenario, we look at both how well the model predicts the rumour's truthfulness and how accurately it identifies the key turning points in the conversation. The results are shown in Table~\ref{tab:turnaround_res_appendix}.

\section{Labelled Samples in PHEME}
\label{sec: labelled_samples_appendix}
\begin{figure}[t]
\centering
  \includegraphics[width=\linewidth]{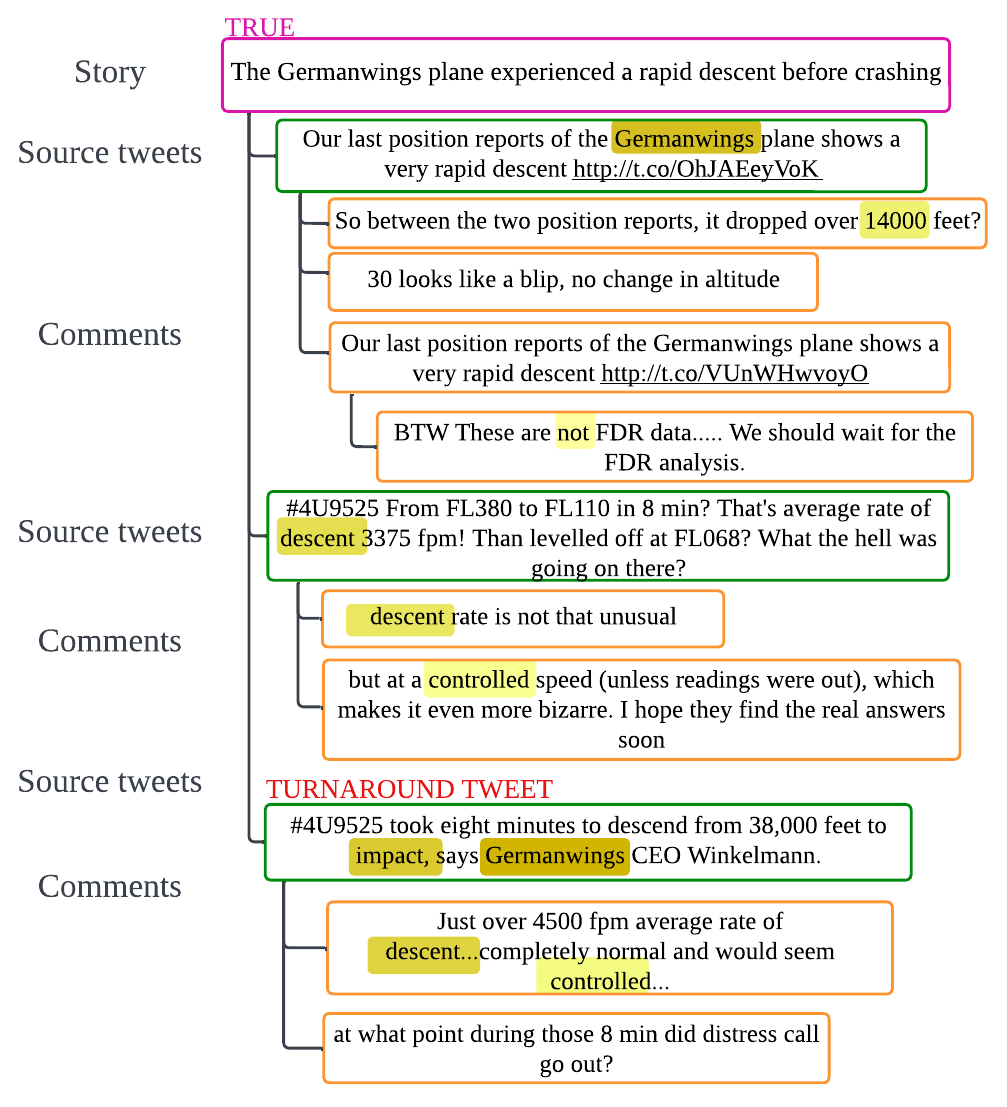}
  \caption{A labelled true story in PHEME. }
  \label{fig:story-pheme-true}
\end{figure}

\begin{figure}[t]
\centering
  \includegraphics[width=\linewidth]{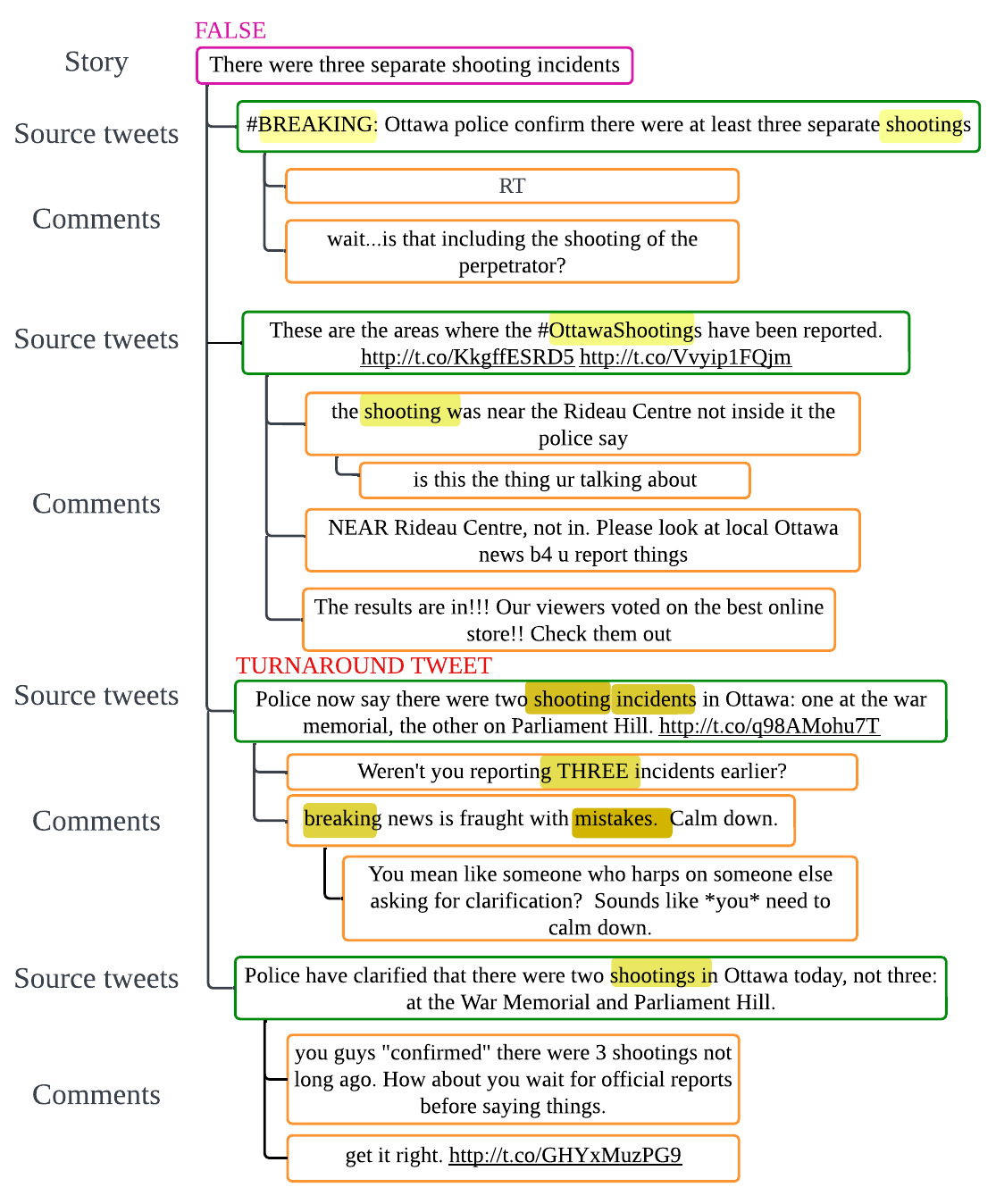}
  \caption{A labelled false story in PHEME. }
  \label{fig:story-pheme-false}
\end{figure}

In order to provide a better understanding of the dataset utilised in our experiments, this section will further include labelled story samples (Figure~\ref{fig:story-pheme-true} and Figure~\ref{fig:story-pheme-false}), supplementing the example presented in Figure~\ref{fig:story-pheme} of the main manuscript, ensuring consistency of our findings.

\section{Hyper-parameter Details}
\label{sec:hyperparameters_appendix}
To fine-tune the base rumour detection model, we use the development set of the dataset for tuning 
hyper-parameters for each model. The detailed searched hyper-parameters are listed in Table~\ref{tab:hyperparameter_table}. 
\begin{table*}[t]
\begin{center}
\begin{adjustbox}{max width=\linewidth}
\begin{tabular}{lcccc}
\toprule
\toprule
& Model & Base Encoder & Learning Rate & Dropout Rate \\ \toprule
& One-tier Transformer & RoBERTa & [3e-5, 5e-5] & [0.4-0.5] \\
& Two-tier Transformer & BERT & [2e-5,5e-5] & [0.5-0.6] \\
& DUCK & BERT & [1e-5, 5e-5] & [0.1-0.2] \\
\bottomrule
\bottomrule 
\end{tabular}
\end{adjustbox}
\end{center}
\caption{\label{tab:hyperparameter_table} Hyper-parameters.}
\end{table*}

\begin{table*}[t]
\begin{center}
\begin{adjustbox}{max width=\linewidth}
\begin{tabular}{lcccc}
\toprule
\toprule
& Dataset & \# source tweet & \#comments & \# stories \\ \toprule
& PHEME~\cite{kochkina2018all} & 6,245 & 98,929 & -- \\
& PHEME~\cite{zubiaga2016analysing} & 7,507 & 32,154 & 240 \\
\bottomrule
\bottomrule 
\end{tabular}
\end{adjustbox}
\end{center}
\caption{\label{tab:dataset_table} Datasets Statistics.}
\end{table*}

\section{Statistical Test}
\label{sec:statisticaltest_appendix}
In the qualitative analysis, we conducted significance tests to validate the performance improvements across three types of interpretability models.
We conducted Man-Whitney tests on accuracy for identifying turnaround posts. Results show that CMA-R is statistically significantly better than other interpretability models \( p-value \ll 0.05 \). Results are shown in Table~\ref{tab:ttest_analysis}.

\begin{table*}[t]
\begin{center}
\begin{adjustbox}{max width=\linewidth}
\begin{tabular}{rcccc}
\toprule
\toprule
& Model & Pairs  & P-value \\ \toprule
& One-Tier & CMA-R vs Random  & 0.00016 \\
& One-Tier & CMA-R vs Attention & 0.00348 \\
& One-Tier & CMA-R vs Local  & 0.00138 \\
& One-Tier & CMA-R vs Gradient  & 0.02925 \\
& Two-Tier & CMA-R vs Random  & 0.00015 \\
& Two-Tier & CMA-R vs Attention  & 0.00040 \\
& Two-Tier & CMA-R vs Local  & 0.00055 \\
& Two-Tier & CMA-R vs Gradient  & 0.01040 \\
& DUCK & CMA-R vs Random  & 0.00016 \\
& DUCK & CMA-R vs Attention  & 0.00040 \\
& DUCK & CMA-R vs Local  & 0.00040 \\
& DUCK & CMA-R vs Gradient  & 0.00467 \\
\bottomrule
\bottomrule 
\end{tabular}
\end{adjustbox}
\end{center}
\caption{\label{tab:ttest_analysis} Mann-Whitney U test results.}
\end{table*}






\end{document}